# Smoothing Proximal Gradient Method for General Structured Sparse Learning

Xi Chen[1], Qihang Lin[2], Seyoung Kim[1], Jaime G. Carbonell[1] and Eric P. Xing[1]

[1] School of Computer Science, Carnegie Mellon University, Pittsburgh, PA, 15213, USA
{xichen,sssykim,jgc,epxing}@cs.cmu.edu

[2] Tepper School of Business, Carnegie Mellon University, Pittsburgh, PA, 15213, USA
{qihangl}@andrew.cmu.edu

## Abstract

We study the problem of learning high dimensional regression models regularized by a structured-sparsity-inducing penalty that encodes prior structural information on either input or output sides. We consider two widely adopted types of such penalties as our motivating examples: 1) overlapping group lasso penalty, based on the $\ell_1/\ell_2$ mixed-norm penalty, and 2) graph-guided fusion penalty. For both types of penalties, due to their non-separability, developing an efficient optimization method has remained a challenging problem. In this paper, we propose a general optimization approach, called smoothing proximal gradient method, which can solve the structured sparse regression problems with a smooth convex loss and a wide spectrum of structured-sparsity-inducing penalties. Our approach is based on a general smoothing technique of Nesterov. It achieves a convergence rate faster than the standard first-order method, subgradient method, and is much more scalable than the most widely used interior-point method. Numerical results are reported to demonstrate the efficiency and scalability of the proposed method.

## 1 Introduction

While $\ell_1$-regularized regression (e.g., lasso [21]) is widely used for variable selection in high-dimensional space, it is not capable of capturing any structural information among variables. In recent years, a technique known as *regularization with structured-sparsity-inducing penalty* has been employed to take advantage of prior knowledge of the structures among inputs (or outputs), to encourage closely related inputs to be selected jointly [24, 22, 6, 9, 8]. Although intricate structured-sparsity-inducing penalties are now widely adopted by both modelers and practitioners, developing efficient optimization algorithms to solve the resultant estimation problem under general class of such penalties remains a significant challenge.

When the structure to be imposed has a relatively simple form, such as non-overlapping groups over variables (e.g., group lasso [24]), or a linear-ordering (a.k.a., chain structure) of variables (e.g., fused lasso [22]), efficient optimization methods have been developed. For example, under group lasso [24], due to the separability among groups, a *proximal operator*[1] associated with the penalty can be computed in a closed-form; thus, a number of composite gradient methods [2, 16, 11] that leverage the proximal operator as a key step (so-called "proximal gradient method") can be directly applied. For fused lasso, although the penalty is not separable, a coordinate descent algorithm was shown feasible by explicitly leveraging the linear ordering of the input variables [5].

In order to handle a more general class of structures such as tree or graph, various methods that further extend group lasso and fused lasso have been proposed. While the standard group lasso assumes groups are not overlapping, *overlapping group lasso* [6] allows us to incorporate more complex prior knowledge on the structure by introducing overlaps among groups so that each input can belong to multiple groups. As for fused lasso, *graph-guided fused lasso* extends the chain structure to a general graph, where the fusion penalty is applied to each edge of the graph [8]. However, due to the non-separability of the penalty that arises from overlapping groups or graphs, the fast optimization method for the standard group lasso or fused lasso cannot be easily applied (e.g., no closed-form solution of the proximal operator). In principle, generic solvers such as the interior-point methods (IPM) could always be used to solve either a second-order cone programming (SOCP) or a quadratic programming (QP) formulation of the aforementioned problems, such approaches are computationally expensive even for problems of moderate size. Very recently, this problem has received a great deal of attention from a number of papers [7, 15, 4, 14, 12, 13] which all strived to provide clever solutions to various subclasses of the structured-sparsity-inducing penalties. However, as we survey in Section 4, they are still shy of reaching a simple, unified, and general solution to a broad class of structured sparse learning problems.

---

[1] The proximal operator associated with the penalty is defined as $\arg\min_{\boldsymbol{\beta}} \frac{1}{2}\|\boldsymbol{\beta}-\mathbf{v}\|_2^2 + P(\boldsymbol{\beta})$, where $\mathbf{v}$ is any given vector and $P(\boldsymbol{\beta})$ is the non-smooth penalty.

In this paper, we propose a generic optimization approach, the smoothing proximal gradient method, for dealing with a variety of structured-sparsity-inducing penalties. We use overlapping group lasso penalty and graph-guided fusion penalty as our motivating examples. Although these two types of penalties are seemingly very different, we show that it is possible to decouple the non-separable terms in both penalties via the dual norm; and reformulate them into a common form to which the proposed method can be applied. We call this approach "smoothing" proximal gradient method because instead of optimizing the original problem directly as in other proximal gradient methods, we introduce a *smooth* approximation of the structured-sparsity-inducing penalty using the Nesterov's smoothing technique [17]. Then, we solve the approximation problem by the first-order proximal gradient method: fast iterative shrinkage-thresholding algorithm (FISTA)[2]. It achieves $O(\frac{1}{\epsilon})$ convergence rate for a desired accuracy $\epsilon$. Below, we summarize the main advantages of this approach.

(a) It is a first-order method, as it uses only the gradient information. Thus, it is significantly more scalable than IPM for SOCP or QP. Since it is gradient-based, it allows warm restarts, thereby potentiates solving the problem along the entire regularization path [5].

(b) It enjoys a convergence rate of $O(\frac{1}{\epsilon})$, which dominates that of the standard subgradient method with rate $O(\frac{1}{\epsilon^2})$.

(c) It is applicable to a wide class of optimization problems with a smooth convex loss and a non-smooth non-separable structured-sparsity-inducing penalty. In particular, it is applicable to both uni- and multi-task learning, with structures on either (or both) inputs/outputs.

(d) Easy to implement with only a few lines of MATLAB code.

## 2 Preliminary

We begin with a review of the high-dimensional linear regression model regularized by structured-sparsity-inducing penalties.

Given the input data $\mathbf{X} \in \mathbb{R}^{N \times J}$ for $N$ samples where each sample lies in $J$ dimensional space; and the output data $\mathbf{y} \in \mathbb{R}^{N \times 1}$, we assume a linear regression model, $\mathbf{y} = \mathbf{X}\boldsymbol{\beta} + \boldsymbol{\epsilon}$, where $\boldsymbol{\beta}$ is the vector of regression coefficients and $\boldsymbol{\epsilon}$ is the noise distributed as $N(0, \sigma^2 I_{N \times N})$. Lasso [21] finds a sparse estimate of the parameters by optimizing:

$$\min_{\boldsymbol{\beta} \in \mathbb{R}^J} g(\boldsymbol{\beta}) + \lambda \|\boldsymbol{\beta}\|_1, \quad (1)$$

where $g(\boldsymbol{\beta}) \equiv \frac{1}{2}\|\mathbf{y} - \mathbf{X}\boldsymbol{\beta}\|_2^2$ is the squared-error loss; $\|\boldsymbol{\beta}\|_1 \equiv \sum_{j=1}^{J} |\beta_j|$ is the $\ell_1$-norm penalty that enforces the individual feature-level sparsity; and $\lambda$ is the regularization parameter.

If the structure on the inputs is available and related inputs are believed to be jointly relevant or irrelevant, we can incorporate this structural information by introducing a *structured-sparsity-inducing penalty* $\Omega(\boldsymbol{\beta})$ as follows:

$$\min_{\boldsymbol{\beta} \in \mathbb{R}^J} f(\boldsymbol{\beta}) \equiv g(\boldsymbol{\beta}) + \Omega(\boldsymbol{\beta}) + \lambda \|\boldsymbol{\beta}\|_1. \quad (2)$$

In this paper, we consider the following two categories of $\Omega(\boldsymbol{\beta})$ that cover a board set of penalties in the literature [24, 6, 9, 25, 22, 8].

**[1] Overlapping Group Lasso Penalty**: Assuming that the set of groups of inputs $\mathcal{G} = \{g_1, \ldots, g_{|\mathcal{G}|}\}$ is defined as a subset of the power set of $\{1, \ldots, J\}$, and is available as prior knowledge. Note that members of $\mathcal{G}$ (groups) are allowed to overlap. The overlapping group lasso penalty based on the $\ell_1/\ell_2$ mixed-norm [6] is defined as:

$$\Omega(\boldsymbol{\beta}) \equiv \gamma \sum_{g \in \mathcal{G}} w_g \|\boldsymbol{\beta}_g\|_2, \quad (3)$$

here $\boldsymbol{\beta}_g \in \mathbb{R}^{|g|}$ is the subvector of $\boldsymbol{\beta}$ for the inputs in group $g$; $\gamma$ is the regularization parameter for structured sparsity; $w_g$ is the predefined weight for group $g$; and $\|\cdot\|_2$ is the vector $\ell_2$-norm. The $\ell_1/\ell_2$ mixed-norm penalty $\Omega(\boldsymbol{\beta})$ plays the role of setting all of the coefficients within each group to zero or non-zero values. We note that many of the structured-sparsity-inducing penalties in the current literature are a special case of (3). Examples include tree-structured penalty [25, 9], where groups are defined for subtrees at each internal node, and graph-structured penalty, where each group is defined as two nodes of an edge.

**[2] Graph-guided Fusion Penalty**: Now we assume the structure of $J$ inputs is available as a graph $G$ with

a set of nodes $V = \{1, \ldots, J\}$ and edges $E$. The graph-guided fusion penalty is defined as:

$$\Omega(\boldsymbol{\beta}) = \gamma \sum_{e=(m,l) \in E, m<l} |\beta_m - \beta_l|. \quad (4)$$

This penalty achieves the effect that coefficients corresponding to two nodes of an edge tend to be the same.

As a generalization of (4), one can construct the graph by computing pairwise correlation based on $\mathbf{x}_j$'s, and connect two nodes with an edge if their correlation is above a given threshold $\rho$. Let $r_{ml} \in \mathbb{R}$ denote the weight (can be either positive or negative) of an edge $e = (m,l) \in E$ corresponding to the correlation between the two nodes. The generalized graph-guided fusion penalty [8] can be defined as:

$$\Omega(\boldsymbol{\beta}) = \gamma \sum_{e=(m,l) \in E, m<l} \tau(r_{ml}) |\beta_m - \text{sign}(r_{ml})\beta_l|, \quad (5)$$

where $\tau(r)$ weights the fusion penalty for each edge such that $\beta_m$ and $\beta_l$ for highly correlated inputs with larger $|r_{ml}|$ receive a greater fusion effect. In this paper, we consider $\tau(r) = |r|$. The $\text{sign}(r_{ml})$ indicates that for two positively correlated nodes, the corresponding coefficients tend to be influence the output in the same direction, while for two negatively correlated nodes, the effects ($\beta_m$ and $\beta_l$) take the opposite direction. If $r_{ml} = 1$ for all $e = (m,l)$, the penalty in (5) reduces to the simple form in (4).

## 3 Smoothing Proximal Gradient

In this section, we present the smoothing proximal gradient method. The main difficulty in optimizing (2) arises from the non-separability of $\boldsymbol{\beta}$ in the non-smooth penalty $\Omega(\boldsymbol{\beta})$. For both types of penalties, we show that using the dual norm, the non-separable structured-sparsity-inducing penalties $\Omega(\boldsymbol{\beta})$ can be formulated as $\Omega(\boldsymbol{\beta}) = \max_{\boldsymbol{\alpha} \in \mathcal{Q}} \boldsymbol{\alpha}^T C \boldsymbol{\beta}$. Based on that, we introduce the smooth approximation to $\Omega(\boldsymbol{\beta})$ using a general smoothing technique of Nesterov [17].

### 3.1 Reformulation of the Penalty

**[1] Overlapping Group Lasso Penalty** Since the dual norm of $\ell_2$-norm is also an $\ell_2$-norm, we can write $\|\boldsymbol{\beta}_g\|_2$ as $\|\boldsymbol{\beta}_g\|_2 = \max_{\|\boldsymbol{\alpha}_g\|_2 \leq 1} \boldsymbol{\alpha}_g^T \boldsymbol{\beta}_g$, where $\boldsymbol{\alpha}_g \in \mathbb{R}^{|g|}$ is the vector of auxiliary variables associated with $\boldsymbol{\beta}_g$. Let $\boldsymbol{\alpha} = \left[\boldsymbol{\alpha}_{g_1}^T, \ldots, \boldsymbol{\alpha}_{g_{|\mathcal{G}|}}^T\right]^T$ with its domain $\mathcal{Q} \equiv \{\boldsymbol{\alpha} \mid \|\boldsymbol{\alpha}_g\|_2 \leq 1, \forall g \in \mathcal{G}\}$, where $\mathcal{Q}$ is the product of unit $\ell_2$ balls. We can rewrite overlapping group lasso penalty (3) as:

$$\begin{aligned} \Omega(\boldsymbol{\beta}) &= \gamma \sum_{g \in \mathcal{G}} w_g \|\boldsymbol{\beta}_g\|_2 = \gamma \sum_{g \in \mathcal{G}} w_g \max_{\|\boldsymbol{\alpha}_g\|_2 \leq 1} \boldsymbol{\alpha}_g^T \boldsymbol{\beta}_g \\ &= \max_{\boldsymbol{\alpha} \in \mathcal{Q}} \sum_{g \in \mathcal{G}} \gamma w_g \boldsymbol{\alpha}_g^T \boldsymbol{\beta}_g = \max_{\boldsymbol{\alpha} \in \mathcal{Q}} \boldsymbol{\alpha}^T C \boldsymbol{\beta}, \quad (6) \end{aligned}$$

where $C \in \mathbb{R}^{\sum_{g \in \mathcal{G}} |g| \times J}$ is a matrix defined as follows. The rows of $C$ are indexed by all pairs of $(i,g) \in \{(i,g) | i \in g, i \in \{1, \ldots, J\}\}$, the columns are indexed by $j \in \{1, \ldots, J\}$, and each element of $C$ is given as:

$$C_{(i,g),j} = \begin{cases} \gamma w_g & \text{if } i = j, \\ 0 & \text{otherwise.} \end{cases}$$

Note that $C$ is a highly sparse matrix with only a single non-zero element in each row and $\sum_{g \in \mathcal{G}} |g|$ non-zero elements in the entire matrix, and hence, can be stored with only a small amount of memory during the optimization procedure.

**[2] Graph-guided Fusion Penalty** First, we rewrite the graph-guided fusion penalty in (5) as follows:

$$\gamma \sum_{e=(m,l) \in E, m<l} \tau(r_{ml}) |\beta_m - \text{sign}(r_{ml})\beta_l| \equiv \|C\boldsymbol{\beta}\|_1,$$

where $C \in \mathbb{R}^{|E| \times J}$ is the edge-vertex incident matrix defined as below:

$$C_{e=(m,l),j} = \begin{cases} \gamma \cdot \tau(r_{ml}) & \text{if } j = m \\ -\gamma \cdot \text{sign}(r_{ml})\tau(r_{ml}) & \text{if } j = l \\ 0 & \text{otherwise.} \end{cases}$$

Since the $\ell_\infty$-norm and the $\ell_1$-norm are dual of each other, we can further rewrite the graph-guided fusion penalty as:

$$\|C\boldsymbol{\beta}\|_1 \equiv \max_{\|\boldsymbol{\alpha}\|_\infty \leq 1} \boldsymbol{\alpha}^T C \boldsymbol{\beta}, \quad (7)$$

where $\boldsymbol{\alpha} \in \mathcal{Q} = \{\boldsymbol{\alpha} | \|\boldsymbol{\alpha}\|_\infty \leq 1, \boldsymbol{\alpha} \in \mathbb{R}^{|E|}\}$ is a vector of auxiliary variables, and $\|\cdot\|_\infty$ is the $\ell_\infty$-norm.

**Remark 1.** *As a generalization of graph-guided fusion penalty, the method proposed in this paper can be applied to the $\ell_1$-norm of any linear mapping of $\boldsymbol{\beta}$ (i.e., $\Omega(\boldsymbol{\beta}) = \|C\boldsymbol{\beta}\|_1$ for any given $C$).*

### 3.2 Smooth Approximation of the Penalty

The common formulation of $\Omega(\boldsymbol{\beta})$ (i.e., $\Omega(\boldsymbol{\beta}) = \max_{\boldsymbol{\alpha} \in \mathcal{Q}} \boldsymbol{\alpha}^T C \boldsymbol{\beta}$) is still a non-smooth function of $\boldsymbol{\beta}$, and this makes the optimization challenging. To tackle this problem, using a general smoothing technique of Nesterov [17], we construct a smooth approximation of $\Omega(\boldsymbol{\beta})$ as following:

$$f_\mu(\boldsymbol{\beta}) = \max_{\boldsymbol{\alpha} \in \mathcal{Q}} \left(\boldsymbol{\alpha}^T C \boldsymbol{\beta} - \mu d(\boldsymbol{\alpha})\right), \quad (8)$$

where $\mu$ is the positive smoothness parameter and $d(\boldsymbol{\alpha})$ is defined as $\frac{1}{2}\|\boldsymbol{\alpha}\|_2^2$. The original penalty term can be viewed as $f_\mu(\boldsymbol{\beta})$ with $\mu = 0$. It is easy to see that $f_\mu(\boldsymbol{\beta})$ is a lower bound of $f_0(\boldsymbol{\beta})$. In order to bound the gap between $f_\mu(\boldsymbol{\beta})$ and $f_0(\boldsymbol{\beta})$, let $D = \max_{\boldsymbol{\alpha} \in \mathcal{Q}} d(\boldsymbol{\alpha})$. In our problems, $D = |\mathcal{G}|/2$ for the overlapping group lasso penalty and $D = |E|/2$ for the graph-guided fusion penalty. Then, it is easy to verify that the maximum gap is $\mu D$: $f_0(\boldsymbol{\beta}) - \mu D \leq f_\mu(\boldsymbol{\beta}) \leq f_0(\boldsymbol{\beta})$. From Theorem 1 as presented below, we know that $f_\mu(\boldsymbol{\beta})$ is a smooth function for any $\mu > 0$. Therefore, $f_\mu(\boldsymbol{\beta})$ can be viewed as a *smooth approximation* of $f_0(\boldsymbol{\beta})$ with the maximum gap of $\mu D$, and the $\mu$ controls the gap between $f_\mu(\boldsymbol{\beta})$ and $f_0(\boldsymbol{\beta})$. Given the desired accuracy $\epsilon$, the convergence result in Section 3.4 suggests $\mu = \frac{\epsilon}{2D}$ to achieve the best convergence rate.

Now we present the key theorem from [17] to show that $f_\mu(\boldsymbol{\beta})$ is smooth in $\boldsymbol{\beta}$ with a simple form of the gradient.

**Theorem 1.** *For any $\mu > 0$, $f_\mu(\boldsymbol{\beta})$ is convex and continuously-differentiable with the gradient*

$$\nabla f_\mu(\boldsymbol{\beta}) = C^T \boldsymbol{\alpha}^*, \qquad (9)$$

*where $\boldsymbol{\alpha}^*$ is the optimal solution to (8). Moreover, the gradient $\nabla f_\mu(\boldsymbol{\beta})$ is Lipschitz continuous with the Lipschitz constant $L_\mu = \frac{1}{\mu}\|C\|^2$, where $\|C\|$ is the matrix spectral norm of $C$ defined as: $\|C\| \equiv \max_{\|\mathbf{v}\|_2 \leq 1} \|C\mathbf{v}\|_2$.*

By viewing $f_\mu(\boldsymbol{\beta})$ as the *Fenchel Conjugate* of $d(\cdot)$ at $\frac{C\boldsymbol{\beta}}{\mu}$, the smoothness can obtained by applying Theorem 26.3 in [19]. The gradient in (9) can be derived from the Danskin's Theorem [3] and the Lipschitz constant is shown in [17].

**Geometric illustration of Theorem 1** To provide insights on why $f_\mu(\boldsymbol{\beta})$ is smooth function as Theorem 1 suggests, in Figure 1, we show a geometric illustration for the case of one-dimensional parameter (i.e., $\beta \in \mathbb{R}$) with $\mu$ and $C$ set to 1. First, we show geometrically that $f_0(\beta) = \max_{\alpha \in [-1,1]} z(\alpha, \beta)$ with $z(\alpha, \beta) \equiv \alpha\beta$ is a non-smooth function. The three-dimensional plot for $z(\alpha, \beta)$ with $\alpha$ restricted to $[-1, 1]$ is shown in Figure 1(a). We project the surface in Figure 1(a) onto the $\beta - z$ space as shown in Figure 1(b). For each $\beta$, the value of $f_0(\beta)$ is the highest point along the $z$-axis since we maximize over $\alpha$ in $[-1, 1]$. We can see that $f_0(\beta)$ is composed of two segments with a sharp point at $\beta = 0$ and hence is *non-smooth*. Now, we introduce $d(\alpha) = \frac{1}{2}\alpha^2$, let $z_s(\alpha, \beta) \equiv \alpha\beta - \frac{1}{2}\alpha^2$ and $f_\mu(\beta) = \max_{\alpha \in [-1,1]} z_s(\alpha, \beta)$. The three-dimensional plot for $z_s(\alpha, \beta)$ with $\alpha$ restricted to $[-1, 1]$ is shown in Figure 1(c). Similarly, we project the surface in Figure 1(c) onto the $\beta - z_s$ space as shown in Figure 1(d). For fixed $\beta$, the value of $f_\mu(\beta)$ is the highest point along the $z$-axis. In Figure 1(d), we can see that the sharp point at $\beta = 0$ is removed and $f_\mu(\beta)$ becomes *smooth*.

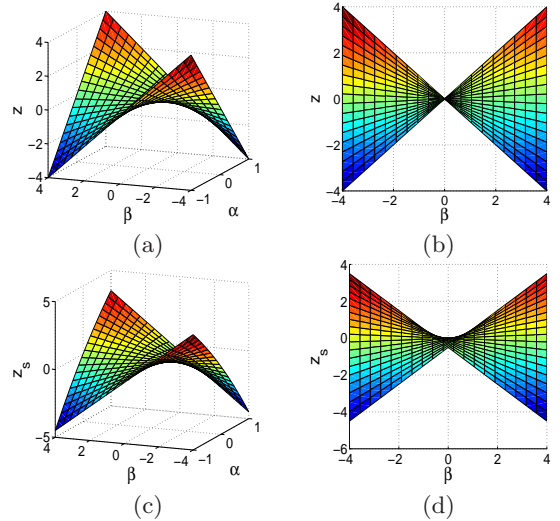

Figure 1: A geometric illustration of the smoothness of $f_\mu(\boldsymbol{\beta})$. (a) The 3-D plot of $z(\alpha, \beta)$, (b) the projection of (a) onto the $\beta$-$z$ space, (c) the 3-D plot of $z_s(\alpha, \beta)$, and (d) the projection of (c) onto the $\beta$-$z$ space.

To compute the $\nabla f_\mu(\boldsymbol{\beta})$ and $L_\mu$, we need to know $\boldsymbol{\alpha}^*$ and $\|C\|$. We present the closed-form equations for $\boldsymbol{\alpha}^*$ and $\|C\|$ for overlapping group lasso and graph-guided fusion penalties in the following propositions.

**[1] Overlapping Group Lasso Penalty**

**Proposition 1.** *Let $\boldsymbol{\alpha}^*$, which is composed of $\{\boldsymbol{\alpha}_g^*\}_{g \in \mathcal{G}}$, be the optimal solution to (8) for overlapping group lasso penalty in (3). For any $g \in \mathcal{G}$,*

$$\boldsymbol{\alpha}_g^* = S_2\Big(\frac{\gamma w_g \boldsymbol{\beta}_g}{\mu}\Big), \qquad (10)$$

*where $S_2$ is the projection operator which projects any vector $\mathbf{u}$ to the $\ell_2$ ball:*

$$S_2(\mathbf{u}) = \begin{cases} \frac{\mathbf{u}}{\|\mathbf{u}\|_2} & \|\mathbf{u}\|_2 > 1, \\ \mathbf{u} & \|\mathbf{u}\|_2 \leq 1. \end{cases}$$

$$\|C\| = \gamma \max_{j \in \{1,\ldots,J\}} \sqrt{\sum_{g \in \mathcal{G} \ s.t. \ j \in g} (w_g)^2}.$$

**[2] Graph-guided Fusion Penalty**

**Proposition 2.** *Let $\boldsymbol{\alpha}^*$ be the optimal solution of (8) for graph-guided fusion penalty in (5). Then, we have:*

$$\boldsymbol{\alpha}^* = S_\infty\Big(\frac{C\boldsymbol{\beta}}{\mu}\Big), \qquad (11)$$

*where $S_\infty$ is the projection operator which projects a number $x$ to the $\ell_\infty$-ball:*

$$S_\infty(x) = \begin{cases} x, & \text{if} \quad -1 \leq x \leq 1 \\ 1, & \text{if} \quad x > 1 \\ -1, & \text{if} \quad x < -1. \end{cases}$$

*For any vector $\boldsymbol{\alpha}$, $S(\boldsymbol{\alpha})$ is defined as applying $S$ on each and every entry of $\boldsymbol{\alpha}$.*

$\|C\|$ is upper-bounded by $\sqrt{2\gamma^2 \max_{j\in V} d_j}$, where

$$d_j = \sum_{e\in E \text{ s.t. } e \text{ incident on } j} (\tau(r_e))^2 \quad (12)$$

for $j \in V$ in graph $G$, and this bound is tight. Note that when $\tau(r_e) = 1$ for all $e \in E$, $d_j$ is simply the degree of the node $j$.

### 3.3 Smoothing Proximal Gradient Method

Given the smooth approximation of the penalty in (8), we now apply the fast iterative shrinkage thresholding algorithm (FISTA) [2] to solve the regression problem regularized by the structured-sparsity-inducing penalty in (2). We substitute the penalty function $\Omega(\boldsymbol{\beta})$ in (2) with its smooth approximation $f_\mu(\boldsymbol{\beta})$ and obtain the optimization problem:

$$\min_{\boldsymbol{\beta}} \widetilde{f}(\boldsymbol{\beta}) \equiv g(\boldsymbol{\beta}) + f_\mu(\boldsymbol{\beta}) + \lambda\|\boldsymbol{\beta}\|_1. \quad (13)$$

Let $h(\boldsymbol{\beta}) = g(\boldsymbol{\beta}) + f_\mu(\boldsymbol{\beta}) = \frac{1}{2}\|\mathbf{y} - \mathbf{X}\boldsymbol{\beta}\|_2^2 + f_\mu(\boldsymbol{\beta})$. According to Theorem 1, the gradient of $h(\boldsymbol{\beta})$ is:

$$\nabla h(\boldsymbol{\beta}) = \mathbf{X}^T(\mathbf{X}\boldsymbol{\beta} - \mathbf{y}) + C^T\boldsymbol{\alpha}^*, \quad (14)$$

Moreover, $\nabla h(\boldsymbol{\beta})$ Lipschitz continuous with the Lipschitz constant: $L = \lambda_{\max}(\mathbf{X}^T\mathbf{X}) + L_\mu = \lambda_{\max}(\mathbf{X}^T\mathbf{X}) + \frac{\|C\|^2}{\mu}$, where $\lambda_{\max}(\mathbf{X}^T\mathbf{X})$ is the largest eigenvalue of $(\mathbf{X}^T\mathbf{X})$.

Since $\widetilde{f}(\boldsymbol{\beta})$ only involves a very simple non-smooth term (i.e., the $\ell_1$-norm penalty), we can adopt FISTA [2], to minimize $\widetilde{f}(\boldsymbol{\beta})$ as shown in Algorithm 1.

---

**Algorithm 1** Smoothing Proximal Gradient Method

**Input**: $\mathbf{X}$, $\mathbf{y}$, $C$, $\boldsymbol{\beta}^0$, desired accuracy $\epsilon$.
**Initialization**: set $\mu = \frac{\epsilon}{2D}$, $\theta_0 = 1$, $\mathbf{w}^0 = \boldsymbol{\beta}^0$.
**Iterate** For $t = 0, 1, 2, \ldots$, until convergence of $\boldsymbol{\beta}^t$:

1. Compute $\nabla h(\mathbf{w}^t) = \mathbf{X}^T(\mathbf{X}\mathbf{w}^t - \mathbf{y}) + C^T\boldsymbol{\alpha}^*$.
2. Solve the proximal operator associated with the $\ell_1$-norm penalty:

$$\begin{aligned}\boldsymbol{\beta}^{t+1} = \quad &\arg\min_{\boldsymbol{\beta}} h(\mathbf{w}^t) + \langle \boldsymbol{\beta} - \mathbf{w}^t, \nabla h(\mathbf{w}^t)\rangle \\ &+ \lambda\|\boldsymbol{\beta}\|_1 + \frac{L}{2}\|\boldsymbol{\beta} - \mathbf{w}^t\|_2^2\end{aligned} \quad (15)$$

3. Set $\theta_{t+1} = \frac{2}{t+3}$.
4. Set $\mathbf{w}^{t+1} = \boldsymbol{\beta}^{t+1} + \frac{1-\theta_t}{\theta_t}\theta_{t+1}(\boldsymbol{\beta}^{t+1} - \boldsymbol{\beta}^t)$.

**Output**: $\widehat{\boldsymbol{\beta}} = \boldsymbol{\beta}^{t+1}$

---

Rewriting (15), we can easily see that it is the proximal operator associated with the $\ell_1$-norm:

$$\boldsymbol{\beta}^{t+1} = \arg\min_{\boldsymbol{\beta}} \frac{1}{2}\|\boldsymbol{\beta} - (\mathbf{w}^t - \frac{1}{L}\nabla h(\mathbf{w}^t))\|_2^2 + \frac{\lambda}{L}\|\boldsymbol{\beta}\|_1.$$

Let $\mathbf{v} = (\mathbf{w}^t - \frac{1}{L}\nabla h(\mathbf{w}^t))$, the closed-form solution of $\boldsymbol{\beta}^{t+1}$ is presented in the next proposition [5].

**Proposition 3.** *The closed form solution to (15) can be obtained by the so-called soft-thresholding operation:*

$$\beta_j = \text{sign}(v_j)\max(0, |v_j| - \frac{\lambda}{L}), \quad j = 1, \ldots, J. \quad (16)$$

A notable advantage of utilizing the proximal operator associated with the simple $\ell_1$-norm penalty is that it can provide us with the exact sparse (zero) solutions due to the soft-thresholding operation.

**Remark 2.** *Algorithm 1 is a general approach that can be applied to any smooth convex loss (e.g., logistic loss) with the structured-sparsity-inducing penalty $\Omega(\boldsymbol{\beta})$ that can be re-written in the form of $\max_{\boldsymbol{\alpha}} \boldsymbol{\alpha}^T C\boldsymbol{\beta}$.*

### 3.4 Convergence Rate and Time Complexity

Although we optimize the approximation function $\widetilde{f}$, it can be proven that the $f(\widehat{\boldsymbol{\beta}})$ is sufficiently close to the optimal objective value of the original function $f(\boldsymbol{\beta}^*)$. We present the convergence rate of Algorithm 1 in the next theorem.

**Theorem 2.** *Let $\boldsymbol{\beta}^*$ be the optimal solution of the original objective function $f$ and $\boldsymbol{\beta}^t$ be the approximate solution at the $t$-th iteration in Algorithm 1. If we require $f(\boldsymbol{\beta}^t) - f(\boldsymbol{\beta}^*) \leq \epsilon$ and set $\mu = \frac{\epsilon}{2D}$, then, the number of iterations $t$ is upper-bounded by*

$$\sqrt{\frac{4\|\boldsymbol{\beta}^* - \boldsymbol{\beta}^0\|_2^2}{\epsilon}\left(\lambda_{\max}(\mathbf{X}^T\mathbf{X}) + \frac{2D\|C\|^2}{\epsilon}\right)}. \quad (17)$$

The convergence rate in (17) can be proven using the proof technique from [10]. More specifically, we can decompose $f(\boldsymbol{\beta}^t) - f(\boldsymbol{\beta}^*)$ into three parts: (i) $f(\boldsymbol{\beta}^t) - \widetilde{f}(\boldsymbol{\beta}^t)$, (ii) $\widetilde{f}(\boldsymbol{\beta}^t) - \widetilde{f}(\boldsymbol{\beta}^*)$, and (iii) $\widetilde{f}(\boldsymbol{\beta}^*) - f(\boldsymbol{\beta}^*)$. (i) and (iii) can be bounded by the gap of the approximation $\mu D$. (ii) only involves $\widetilde{f}$ function and can be upper bounded by $O(\frac{1}{t^2})$ as shown in [2]. We obtain (17) by balancing these three terms. According to Theorem 2, Algorithm 1 converges in $O(\frac{\sqrt{2D}}{\epsilon})$ iterations, which is much faster than the subgradient method with the convergence rate of $O(\frac{1}{\epsilon^2})$. Note that the convergence rate depends on $D$ through the term $\sqrt{2D}$, which again depends on the problem size.

As for the time complexity, assuming that we can pre-compute and store $\mathbf{X}^T\mathbf{X}$ and $\mathbf{X}^T\mathbf{y}$ with the time complexity of $O(J^2 N)$, the main computational cost in each iteration comes from calculating the gradient $\nabla h(\mathbf{w}^t)$. The per-iteration of Algorithm 1 (Prox-Grad) is summarized in Table 1. It shares a similar per-iteration time complexity as the subgradient method but a faster convergence rate. As for the generic solver, IPM for SOCP, although it converges in $\log(\frac{1}{\epsilon})$ iterations, its per-iteration complexity is higher by orders of magnitude than ours as in Table 1.

Table 1: Comparison of Per-iteration Time Complexity

| | Group |
|---|---|
| Prox-Grad | $O(J^2 + \sum_{g \in \mathcal{G}} |g|)$ |
| IPM for SOCP | $O\left((J + |\mathcal{G}|)^2 (N + \sum_{g \in \mathcal{G}} |g|)\right)$ |
| | Graph |
| Prox-Grad | $O(J^2 + |E|)$ |
| IPM for SOCP | $O\left((J + |E|)^2 (N + J + |E|)\right)$ |

**Remark 3.** *If we pre-compute and store $\mathbf{X}^T \mathbf{X}$, the per-iteration time complexity of Algorithm 1 is independent of sample size $N$ as in Table 1. If $J$ is very large, $\mathbf{X}^T \mathbf{X}$ may not fit into memory. In such a case, we compute $\mathbf{X}^T(\mathbf{X}\mathbf{w}^t)$ for each iteration; and the per-iteration complexity will increase by a factor of $N$ but still less than that for IPM for SOCP. Also for logistic loss, the per-iteration complexity is linear in $N$.*

### 3.5 Summary and Discussions

The insight of our work was drawn from two lines of earlier works. The first one is the proximal gradient methods (e.g., Nesterov's composite gradient method [16], FISTA [2]). They have been widely adopted to solve optimization problems with a convex loss and a relatively simple non-smooth penalty, and achieve $O(\frac{1}{\sqrt{\epsilon}})$ convergence rate. However, the complex structure of the non-separable penalties considered in this paper makes it intractable to solve the proximal operator exactly. This is the challenge that we circumvent via smoothing.

The second work, smoothing the non-smooth function, was first proposed in [17]. The algorithm presented in [17] only works for smooth problems so that it has to smooth out the entire non-smooth penalty (including the $\ell_1$-norm). However, it is precisely the *non-smoothness* of the penalty that leads to exact zeros in optimal solutions. Therefore, although widely adopted in optimization field, this approach cannot yield exact zero solution and leads the problem of where to truncate the solution to zero. Moreover, the algorithm in [17] requires the condition that $\boldsymbol{\beta}$ is bounded and that the number of iterations is pre-defined, which may be impractical for real applications. Instead, our approach combines the smoothing technique with the proximal gradient method and hence leads to the exact sparse solutions.

## 4 Related First-order Optimization Methods

First-order method has recently become a popular optimization technique for solving sparse learning problems due to its lower cost per-iteration and good scalability. For the general mixed-norm based group lasso penalties, most of the existing first-order methods can handle only a specific subclass. In particular, most of these methods use the proximal gradient framework [2, 16] and focus on the issue of how to *exactly* solve the proximal operator. For non-overlapping groups with the $\ell_1/\ell_2$ or $\ell_1/\ell_\infty$ mixed-norm penalty, the proximal operator can be solved via a simple projection [11, 4]. A one-pass coordinate ascent method has been developed for tree-structured groups with the $\ell_1/\ell_2$ or $\ell_1/\ell_\infty$ mixed-norm penalty [7, 13], and quadratic min-cost network flow for arbitrary overlapping groups with the $\ell_1/\ell_\infty$ mix-norm penalty [15].

Table 2 summarizes the applicability, the convergence rate, and the per-iteration time complexity for the available first-order methods for different subclasses of the group lasso penalty. As we can see from Table 2, although our method is not the most ideal one for some of the special cases, our method along with FOBOS [4] and the subgradient descent are the generic first-order methods applicable to all subclasses of the penalties. From Table 2, for arbitrary overlaps with the $\ell_1/\ell_\infty$, although the method proposed in [15] achieves $O(\frac{1}{\sqrt{\epsilon}})$ convergence rate, the per-iteration complexity could be high due to solving a quadratic min-cost network flow problem. From the worst-case analysis, the per-iteration time complexity for solving the network flow problem in [15] is at least $O(|V||E|) = O((J + |\mathcal{G}|)(|\mathcal{G}| + J + \sum_{g \in \mathcal{G}} |g|))$, which is much higher than our method with $O(\sum_{g \in \mathcal{G}} |g| \log |g|)$. More importantly, for the case of arbitrary overlaps with the $\ell_1/\ell_2$, our method has a superior convergence rate to all the other methods. In addition, another first-order method [12] was proposed for arbitrary overlapping group lasso which approximately solves the proximal operator. However, since the error introduced in solving each proximal operator will be accumulated over iterations, there is no known convergence result. Another possible approach is the iteratively reweighted least squares [1]. However, its per-iteration time complexity is high due to solving linear systems at each iteration.

For the graph-guided fusion penalty, when the structure is a simple chain, pathwise coordinate descent method [5] can be applied. For the general graph structure, a first-order method that approximately solves the proximal operator was proposed in [14]. Still, there is no known convergence result due to the errors introduced in computing the proximal operator

Table 2: Comparisons of different first-order methods for optimizing mixed-norm based overlapping group lasso penalties. The first column gives either the algorithm for solving the proximal operator in proximal gradient methods (the first three rows) or the optimization approach (the last two rows). Each entry contains the convergence rate and the per-iteration time complexity. For the sake of simplicity, in all of the entries, we omit the time for computing the gradient of the loss function which is needed for all the methods (i.e., $\nabla g(\boldsymbol{\beta})$ with $O(J^2)$). The per-iteration time complexity may come from the computation of the proximal operator or the subgradient of the penalty. "N.A." standards for "not applicable" or no guarantee in the convergence.

| Method | No overlap $\ell_1/\ell_2$ | No overlap $\ell_1/\ell_\infty$ | Overlap Tree $\ell_1/\ell_2$ | Overlap Tree $\ell_1/\ell_\infty$ | Overlap Arbitrary $\ell_1/\ell_2$ | Overlap Arbitrary $\ell_1/\ell_\infty$ |
|---|---|---|---|---|---|---|
| Projection [11] | $O(\frac{1}{\sqrt{\epsilon}})$, $O(J)$ | $O(\frac{1}{\sqrt{\epsilon}})$, $O(J \log J)$ | N.A. | N.A. | N.A. | N.A. |
| Coordinate Ascent [7, 13] | $O(\frac{1}{\sqrt{\epsilon}})$, $O(J)$ | $O(\frac{1}{\sqrt{\epsilon}})$, $O(J \log J)$ | $O(\frac{1}{\sqrt{\epsilon}})$, $O(\sum_{g\in\mathcal{G}} |g|)$ | $O(\frac{1}{\sqrt{\epsilon}})$, $O(\sum_{g\in\mathcal{G}} |g|\log |g|)$ | N.A. | N.A. |
| Network Flow [15] | N.A. | $O(\frac{1}{\sqrt{\epsilon}})$, quadratic min-cost flow | N.A. | $O(\frac{1}{\sqrt{\epsilon}})$, quadratic min-cost flow | N.A. | $O(\frac{1}{\sqrt{\epsilon}})$, quadratic min-cost flow |
| FOBOS [4]/Subgradient | $O(\frac{1}{\epsilon})$, $O(J)$ | $O(\frac{1}{\epsilon})$, $O(J \log J)$ | $O(\frac{1}{\epsilon})$, $O(\sum_{g\in\mathcal{G}} |g|)$ | $O(\frac{1}{\epsilon})$, $O(\sum_{g\in\mathcal{G}} |g|\log |g|)$ | $O(\frac{1}{\epsilon^2})$, $O(\sum_{g\in\mathcal{G}} |g|)$ (subgradient) | $O(\frac{1}{\epsilon})$, quadratic min-cost flow |
| Smoothing Proximal Gradient | $O(\frac{1}{\epsilon})$, $O(J)$ | $O(\frac{1}{\epsilon})$, $O(J \log J)$ | $O(\frac{1}{\epsilon})$, $O(\sum_{g\in\mathcal{G}} |g|)$ | $O(\frac{1}{\epsilon})$, $O(\sum_{g\in\mathcal{G}} |g|\log |g|)$ | $O(\frac{1}{\epsilon})$, $O(\sum_{g\in\mathcal{G}} |g|)$ | $O(\frac{1}{\epsilon})$, $O(\sum_{g\in\mathcal{G}} |g|\log |g|)$ |

over iterations.

## 5 Multi-task Extension

In this section, we show that the smoothing proximal gradient method can be applied in a straightforward manner to multi-task learning setting, where the structural information is available for outputs (tasks). For the sake of simplicity, we discuss case where different tasks share the same input data matrix.

Let $\mathbf{X} \in \mathbb{R}^{N \times J}$ denote the matrix of input data for $J$ inputs over $N$ samples and $\mathbf{Y} \in \mathbb{R}^{N \times K}$ denote the matrix of output data for $K$ outputs. We use a linear model for the $k$-th task: $\mathbf{y}_k = \mathbf{X}\boldsymbol{\beta}_k + \boldsymbol{\epsilon}_k$, $\forall k = 1, \ldots K$, where $\boldsymbol{\beta}_k = [\beta_{1k}, \ldots, \beta_{Jk}]^T$ is the regression-coefficient vector for the $k$-th task and $\boldsymbol{\epsilon}_k$ is Gaussian noise. Let $\mathbf{B} = [\boldsymbol{\beta}_1, \ldots, \boldsymbol{\beta}_K] \in \mathbb{R}^{J \times K}$ be the matrix of regression coefficients for all of the $K$ tasks. Then, the multi-task sparse regression problem can be naturally formulated as the following optimization problem:

$$\min_{\mathbf{B} \in \mathbb{R}^{J \times K}} f(\mathbf{B}) \equiv \frac{1}{2} \|\mathbf{Y} - \mathbf{XB}\|_F^2 + \Omega(\mathbf{B}) + \lambda \|\mathbf{B}\|_1, \quad (18)$$

where $\|\cdot\|_F$ denotes the matrix Frobenius norm, $\|\cdot\|_1$ denotes the matrix entry-wise $\ell_1$-norm and $\Omega(\mathbf{B})$ is a structured-sparsity-inducing penalty with the structural information on the tasks.

**[1] Multi-task Overlapping Group Lasso Penalty**

$$\Omega(\mathbf{B}) \equiv \gamma \sum_{j=1}^{J} \sum_{g \in \mathcal{G}} w_g \|\boldsymbol{\beta}_{jg}\|_2, \quad (19)$$

where $\mathcal{G} = \{g_1, \ldots, g_{|\mathcal{G}|}\}$ is a subset of the power set of $\{1, \ldots, K\}$, and $\boldsymbol{\beta}_{jg}$ is the vector of regression coefficients $\{\beta_{jk}, k \in g\}$. Both $\ell_1/\ell_2$ mixed-norm for multi-task lasso [18] and tree-guided group lasso penalty [9] are special cases of (19).

**[2] Multi-task Graph-guided Fusion Penalty**

$$\Omega(\mathbf{B}) = \gamma \sum_{e=(m,l) \in E} \tau(r_{ml}) \sum_{j=1}^{J} |\boldsymbol{\beta}_{jm} - \text{sign}(r_{ml})\boldsymbol{\beta}_{jl}|, \quad (20)$$

where the structure of the $K$ outputs is available as a graph $G$ with the nodes $V = \{1, \ldots, K\}$ and edges $E$.

Using the similar techniques in Section 3.1, $\Omega(\mathbf{B})$ can be reformulated as:

$$\Omega(\mathbf{B}) = \max_{\mathbf{A} \in \mathcal{Q}} \langle C\mathbf{B}^T, \mathbf{A} \rangle, \quad (21)$$

where $\langle \mathbf{U}, \mathbf{V} \rangle \equiv \text{Tr}(\mathbf{U}^T \mathbf{V})$ denotes a matrix inner product. $C$ is constructed in the similar way just by replacing the index of the input variables by the output variables; and $\mathbf{A}$ is the auxiliary variables.

Then we introduce the smooth approximation of (21):

$$f_\mu(\mathbf{B}) = \max_{\mathbf{A} \in \mathcal{Q}} \langle C\mathbf{B}^T, \mathbf{A} \rangle - \mu d(\mathbf{A}), \quad (22)$$

where $d(\mathbf{A}) \equiv \frac{1}{2} \|\mathbf{A}\|_F^2$. Following a proof strategy similar to that in Theorem 1, we can show that $f_\mu(\mathbf{B})$ is convex and smooth with gradient $\nabla f_\mu(\mathbf{B}) = (\mathbf{A}^*)^T C$, where $\mathbf{A}^*$ is the optimal solution to (22). The closed-form solution of $\mathbf{A}^*$ and the Lipschitz constant for $\nabla f_\mu(\mathbf{B})$ can be derived in the same way.

By substituting $\Omega(\mathbf{B})$ in (18) with $f_\mu(\mathbf{B})$, we can adopt Algorithm 1 to solve (18) with convergence rate of $O(\frac{1}{\epsilon})$. The per-iteration time complexity of our method is $O(J^2 K + J \sum_{g \in \mathcal{G}} |g|)$ for overlapping group lasso and $O(J^2 K + J|E|)$ for graph-guided fused lasso.

Table 3: Comparisons of different optimization methods on the overlapping group lasso

| $|\mathcal{G}| = 10$ ($J = 910$) | | N=1,000 | | N=5,000 | | N=10,000 | |
|---|---|---|---|---|---|---|---|
| | | CPU (s) | Obj. | CPU (s) | Obj. | CPU (s) | Obj. |
| $\gamma = 2$ | SOCP | 103.71 | 266.68 | 493.08 | 917.13 | 3777.46 | 1765.52 |
| | FOBOS | 27.12 | 266.95 | 1.71 | 918.02 | 1.48 | 1765.61 |
| | Prox-Grad | 0.87 | 266.95 | 0.71 | 917.46 | 1.28 | 1765.69 |
| $\gamma = 0.5$ | SOCP | 106.02 | 83.30 | 510.56 | 745.10 | 3585.77 | 1596.42 |
| | FOBOS | 32.44 | 82.99 | 4.98 | 745.79 | 4.65 | 1597.53 |
| | Prox-Grad | 0.42 | 83.39 | 0.41 | 745.10 | 0.69 | 1596.45 |
| $|\mathcal{G}| = 50$ ($J = 4510$) | | N=1,000 | | N=5,000 | | N=10,000 | |
| | | CPU (s) | Obj. | CPU (s) | Obj. | CPU (s) | Obj. |
| $\gamma = 10$ | SOCP | 4144.20 | 1089.01 | - | - | - | - |
| | FOBOS | 476.91 | 1191.05 | 394.75 | 1533.31 | 79.82 | 2263.49 |
| | Prox-Grad | 56.35 | 1089.05 | 77.61 | 1533.32 | 78.90 | 2263.60 |
| $\gamma = 2.5$ | SOCP | 3746.43 | 277.91 | - | - | - | - |
| | FOBOS | 478.62 | 286.33 | 867.94 | 559.25 | 183.72 | 1266.73 |
| | Prox-Grad | 33.09 | 277.94 | 30.13 | 504.34 | 26.74 | 1266.72 |
| $|\mathcal{G}| = 100$ ($J = 9010$) | | N=1,000 | | N=5,000 | | N=10,000 | |
| | | CPU (s) | Obj. | CPU (s) | Obj. | CPU (s) | Obj. |
| $\gamma = 20$ | FOBOS | 1336.72 | 2090.81 | 2261.36 | 3132.13 | 1091.20 | 3278.20 |
| | Prox-Grad | 234.71 | 2090.79 | 225.28 | 2692.98 | 368.52 | 3278.22 |
| $\gamma = 5$ | FOBOS | 1689.69 | 564.21 | 2287.11 | 1302.55 | 3342.61 | 1185.66 |
| | Prox-Grad | 169.61 | 541.61 | 192.92 | 736.56 | 176.72 | 1114.93 |

## 6 Experiment

In this section, we evaluate the scalability and efficiency of the smoothing proximal gradient method (Prox-Grad) on simulated data. For overlapping group lasso, we compare the Prox-Grad with the FOBOS [4] and IPM for SOCP.[2] For graph-guided fused lasso, we compare the running time of Prox-Grad with that of the FOBOS [4] and IPM for QP.[3] Note that for FOBOS, since the proximal operator associated with $\Omega(\boldsymbol{\beta})$ cannot be solved exactly, we set the "loss function" to be $l(\boldsymbol{\beta}) = g(\boldsymbol{\beta}) + \Omega(\boldsymbol{\beta})$ and the penalty to $\lambda \|\boldsymbol{\beta}\|_1$. According to [4], for the non-smooth loss $l(\boldsymbol{\beta})$, FOBOS achieves $O\left(\frac{1}{\epsilon^2}\right)$ convergence rate, which is slower than our method.

All experiments are performed on a standard PC with 4GB RAM and the software is written in MATLAB. The main difficulty in comparisons is a fair stopping criterion. Unlike IPM, Prox-Grad and FOBOS do not generate a dual solution, and therefore, it is hard to compute a primal-dual gap, which is the traditional stopping criterion for IPM. For the fair comparison, we use the following stopping criterion. Since it is well known that IPM usually gives more accurate (i.e., lower) objective, we set the objective obtained from IPM as the optimal objective value and stop the first-order methods when the objective is below 1.001 times the optimal objective value. For large datasets for which IPM cannot be applied, we stop the first-order methods when the relative change in the objective is below $10^{-6}$. In addition, we set the maximum iterations to be 20,000.

---
[2] We use the state-of-the-art MATLAB package SDPT3 [23] for SOCP.
[3] We use the commercial package MOSEK (www.mosek.com) for QP. Graph-guided fused lasso can also be solved by SOCP but it is less efficient than QP.

We constrain the regularization parameters such that $\lambda = \gamma$ and we assume that for each group $g$, $w_g = 1$. As for the smoothing parameter $\mu$, $\mu$ is set to $\frac{\epsilon}{2D}$ according to Theorem 2, where $D$ is determined by the problem size. It is natural that for large-scale problems with large $D$, a larger $\epsilon$ can be adopted without affecting the recovery quality significantly. Therefore, instead of setting $\epsilon$, we directly set $\mu = 10^{-4}$, which provides us reasonably good approximation accuracies for different scales of problems. As for FOBOS, we set the learning rate to $\frac{c}{\sqrt{t}}$ as suggested in [4], where $c$ is tuned to be $\frac{0.1}{\sqrt{NJ}}$ for single-task learning and $\frac{0.1}{\sqrt{NJK}}$ for multi-task learning.

### 6.1 Overlapping Group Lasso

We simulate data for a single-task linear regression with an overlapping group structure as described below. Assuming that the inputs are ordered, we define a sequence of groups of 100 adjacent inputs with an overlap of 10 variables between two successive groups so that $\mathcal{G} = \{\{1, \ldots, 100\}, \{91, \ldots, 190\}, \ldots, \{J-99, \ldots, J\}\}$ with $J = 90|\mathcal{G}| + 10$. We set $\beta_j = (-1)^j \exp(-(j-1)/100)$ for $1 \leq j \leq J$. We sample each element of $\mathbf{X}$ from i.i.d. Gaussian distribution, and generate the output data from $\mathbf{y} = \mathbf{X}\boldsymbol{\beta} + \boldsymbol{\epsilon}$, where $\boldsymbol{\epsilon} \sim N(0, I_{N \times N})$.

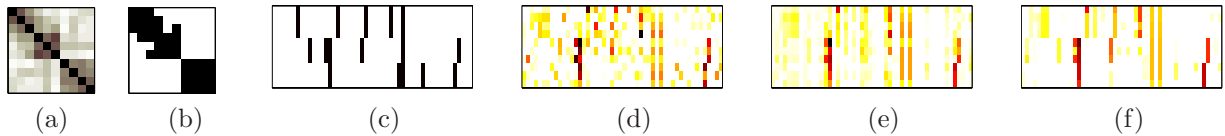

(a) (b) (c) (d) (e) (f)

Figure 2: An example of simulation results. Red pixels indicate large values. (a) The correlation coefficient matrix of outputs. (b) The edges of the output graph obtained at $\rho = 0.3$ are shown as black pixels. (c) The true regression coefficients. Absolute values of the estimated regression coefficients for (d) lasso, (e) $\ell_1/\ell_2$-regularized multi-task lasso, and (f) GFlasso. Rows correspond to outputs and columns to inputs.

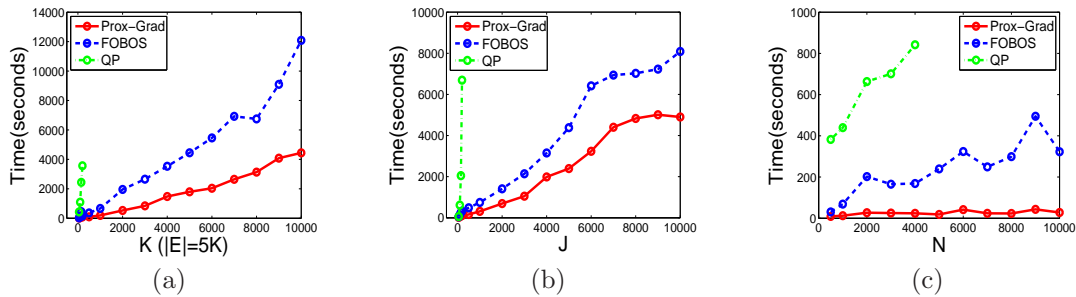

(a) (b) (c)

Figure 3: Comparisons of Prox-Grad, FOBOS and QP. (a) Vary $K$ from 50 to 10,000, fixing $N = 500, J = 100$; (b) Vary $J$ from 50 to 10,000, fixing $N = 1000, K = 50$; and (c) Vary $N$ from 500 to 10000, fixing $J = 100, K = 50$.

To demonstrate the efficiency and scalability of Prox-Grad, we vary $J$, $N$ and $\gamma$ and report the total CPU time in seconds and the objective value in Table 3. The regularization parameter $\gamma$ is set to either $|\mathcal{G}|/5$ or $|\mathcal{G}|/20$. As we can see from Table 3, firstly, both Prox-Grad and FOBOS are more efficient and scalable by orders of magnitude than IPM for SOCP. For larger $J$ and $N$, we are unable to collect the results of SOCP because they lead to out-of-memory errors due to the large storage requirement for solving the Newton linear system. Secondly, Prox-Grad is more efficient than FOBOS for almost all different scales of the problems.[4] Thirdly, for Prox-Grad, a smaller $\gamma$ leads to faster convergence. This result is consistent with Theorem 2 which shows that the number of iterations is linear in $\gamma$ through the term $\|C\|$. Moreover, we notice that a larger $N$ may not increase the computational time for Prox-Grad. This is also consistent with our time complexity analysis, which shows that for linear regression, the per-iteration time complexity is independent of $N$.

### 6.2 Multi-task Graph-guided Fused Lasso

In this section, we apply Prox-Grad method to multi-task graph-guided fused lasso (GFlasso). To show the real applications of GFlasso, we simulate data using the following scenario analogous to genetic association mapping. We use the SNPs in the HapMap CEU panel [20] to simulate the input data $\mathbf{X}$ and set $N = 100$, $J = 30$, $K = 10$. We generate the $\boldsymbol{\beta}_k$'s such that the outputs $\mathbf{y}_k$'s are correlated with a block-like structure in the correlation matrix as in Figure 2(a). We first choose input-output pairs with non-zero regression coefficients as follows. We assume three groups of correlated output variables of sizes 3, 3, and 4 as in Figure 2(c), which correspond to three subgraphs. We randomly select inputs that are relevant jointly to the outputs within each group, and select additional inputs relevant across multiple groups to model the situation of a higher-level correlation structure across two subgraphs. We set all non-zero elements of $\beta_{ij}$'s to a constant $b = 0.8$, and simulate output data based on the linear regression model with noise distributed as $N(0, 1)$. As an illustrative example, the estimated regression coefficients from different methods are shown in Figures 2(d)–(f). While the results of lasso and $\ell_1/\ell_2$-regularized multi-task lasso [18] in Figures 2(d) and (e) contain many false positives, the results from GFlasso in Figure 2(f) show fewer false positives and reveal clear block structures. Thus, GFlasso outperforms the other methods.

---
[4]In some entries in Table 3, the Obj. from FOBOS is much larger than other methods. This is because that FOBOS has reached the maximum number of iterations before convergence.

To compare Prox-Grad with FOBOS and IPM for QP, we vary $K$, $J$, $N$, and present the computation time in seconds in Figures 3(a)-(c), respectively. We select the regularization parameter $\gamma$ using the separate validation dataset, and report the CPU time for GFlasso with the selected $\gamma$. The input, output and true regression coefficient $\mathbf{B}$ are generated in the way similar as above. We choose the $\rho$ for each dataset so that the number of edges is 5 times the number of nodes (i.e. $|E| = 5K$). Figure 3 shows that Prox-Grad is substantially more efficient and can scale up to the datasets with tens of thousands of dimensions or tasks.

# 7 Conclusions

In this paper, we propose a smoothing proximal gradient method for learning a structured-sparsity pattern for a wide spectrum of structured-sparsity-inducing penalties. As for the future work, we observe that setting a large $\mu$ initially and reducing $\mu$ over iterations leads to better empirical results. However, in such a scenario, the convergence rate is harder to analyze. In addition, since the method is only based on gradient, its online version with the stochastic gradient descent can be easily derived. However, proving the regret bound will require more careful investigations.

# 8 Acknowledgements

We would like to thank Javier Peña for the helpful discussion of the related first-order methods. Eric P. Xing is supported by Grants ONR N000140910758, NSF DBI-0640543, NSF CCF-0523757, NIH 1R01GM087694, and an Alfred P. Sloan Research Fellowship.

## References

[1] A. Argyriou, T. Evgeniou, and M. Pontil. Convex multi-task feature learning. *Machine Learning*, 73:243–272, 2008.

[2] A. Beck and M. Teboulle. A fast iterative shrinkage thresholding algorithm for linear inverse problems. *SIAM Journal of Image Science*, 2(1):183–202, 2009.

[3] D. Bertsekas. *Nonlinear Programming*. Athena Scientific, 1999.

[4] J. Duchi and Y. Singer. Efficient online and batch learning using forward backward splitting. *Journal of Machine Learning Research*, 10:2899–2934, 2009.

[5] J. Friedman, T. Hastie, H. Höfling, and R. Tibshirani. Pathwise coordinate optimization. *Annals of Applied Statistics*, 1:302–332, 2007.

[6] R. Jenatton, J. Y. Audibert, and F. Bach. Structured variable selection with sparsity-inducing norms. Technical report, INRIA, 2009.

[7] R. Jenatton, J. Mairal, G. Obozinski, and F. Bach. Proximal methods for sparse hierarchical dictionary learning. In *ICML*, 2010.

[8] S. Kim, K. A. Sohn, and E. P. Xing. A multivariate regression approach to association analysis of a quantitative trait network. *Bioinformatics*, 25(12):204–212, 2009.

[9] S. Kim and E. P. Xing. Tree-guided group lasso for multi-task regression with structured sparsity. In *ICML*, 2010.

[10] G. Lan, Z. Lu, and R. Monteiro. Primal-dual first-order methods with $O(1/\epsilon)$ iteration complexity for cone programming. *Mathematical Programming*, 126:1–29, 2011.

[11] J. Liu, S. Ji, and J. Ye. Multi-task feature learning via efficient $\ell_{2,1}$-norm minimization. In *UAI*, 2009.

[12] J. Liu and J. Ye. Fast overlapping group lasso. ArXiv:1009.0306v1 [cs.LG].

[13] J. Liu and J. Ye. Moreau-yosida regularization for grouped tree structure learning. In *NIPS*, 2010.

[14] J. Liu, L. Yuan, and J. Ye. An efficient algorithm for a class of fused lasso problems. In *ACM SIGKDD*, 2010.

[15] J. Mairal, R. Jenatton, G. Obozinski, and F. Bach. Network flow algorithms for structured spasity. In *NIPS*, 2010.

[16] Y. Nesterov. Gradient methods for minimizing composite objective function. ECORE Discussion Paper 2007.

[17] Y. Nesterov. Smooth minimization of non-smooth functions. *Mathematical Programming*, 103(1):127–152, 2005.

[18] G. Obozinski, B. Taskar, and M. I. Jordan. High-dimensional union support recovery in multivariate regression. In *NIPS*, 2009.

[19] R. Rockafellar. *Convex Analysis*. Princeton Univ. Press, 1996.

[20] The International HapMap Consortium. A haplotype map of the human genome. *Nature*, 437:1399–1320, 2005.

[21] R. Tibshirani. Regression shrinkage and selection via the lasso. *Journal of the Royal Statistical Society: Series B*, 58:267–288, 1996.

[22] R. Tibshirani and M. Saunders. Sparsity and smoothness via the fused lasso. *Journal of the Royal Statistical Society, Series B*, 67(1):91–108, 2005.

[23] R. H. Tütüncü, K. C. Toh, and M. J. Todd. Solving semidefinite-quadratic-linear programs using sdpt3. *Mathematical Programming*, 95:189–217, 2003.

[24] M. Yuan and Y. Lin. Model selection and estimation in regression with grouped variables. *Journal of the Royal Statistical Society: Series B*, 68:49–67, 2006.

[25] P. Zhao, G. Rocha, and B. Yu. The composite absolute penalties family for grouped and hierarchical variable selection. *The Annals of Statistics*, 37(6A):3468–3497, 2009.